\documentclass[runningheads]{llncs}

\usepackage[mobile]{eccv}

\usepackage{eccvabbrv}

\usepackage{graphicx}
\usepackage{booktabs}

\usepackage[accsupp]{axessibility}  % Improves PDF readability for those with disabilities.

\usepackage{hyperref}
\usepackage{multirow}
\usepackage{orcidlink}

\begin{document}

\title{Texture-GS: Disentangling the Geometry and Texture for 3D Gaussian Splatting Editing}

\titlerunning{Texture-GS}

\author{Tian-Xing Xu\inst{1}\and
  Wenbo Hu\inst{2}$^\dag$ \and
  Yu-Kun Lai\inst{3}\and
  Ying Shan\inst{2}\and
  Song-Hai Zhang\inst{1}$^\dag$}

\authorrunning{T.-X. et al.}

\institute{Tsinghua University, China \\
\email{xutx21@mails.tsinghua.edu.cn,shz@tsinghua.edu.cn}\and
 Tencent AI Lab, China \\
\email{wbhu@tencent.com,yingsshan@tencent.com}\and
  Cardiff University, United Kingdom
  \\
  \email{LaiY4@cardiff.ac.uk}}

\maketitle

\renewcommand{\thefootnote}{}
\footnotetext[2]{$^\dag$ Corresponding authors.}

\begin{abstract}
  
  3D Gaussian splatting, emerging as a groundbreaking approach, has drawn increasing attention for its capabilities of high-fidelity reconstruction and real-time rendering.
  However, it couples the appearance and geometry of the scene within the Gaussian attributes, which hinders the flexibility of editing operations, such as texture swapping.
  To address this issue, we propose a novel approach, namely Texture-GS, to disentangle the appearance from the geometry by representing it as a 2D texture mapped onto the 3D surface, thereby facilitating appearance editing.
  Technically, the disentanglement is achieved by our proposed texture mapping module, which consists of a UV mapping MLP to learn the UV coordinates for the 3D Gaussian centers, a local Taylor expansion of the MLP to efficiently approximate the UV coordinates for the ray-Gaussian intersections, and a learnable texture to capture the fine-grained appearance.
  Extensive experiments on the DTU dataset demonstrate that our method not only facilitates high-fidelity appearance editing but also achieves real-time rendering on consumer-level devices, \eg a single RTX 2080 Ti GPU.

\end{abstract}

\section{Introduction}

\begin{figure}[tb]
  \centering
  \includegraphics[width=\linewidth]{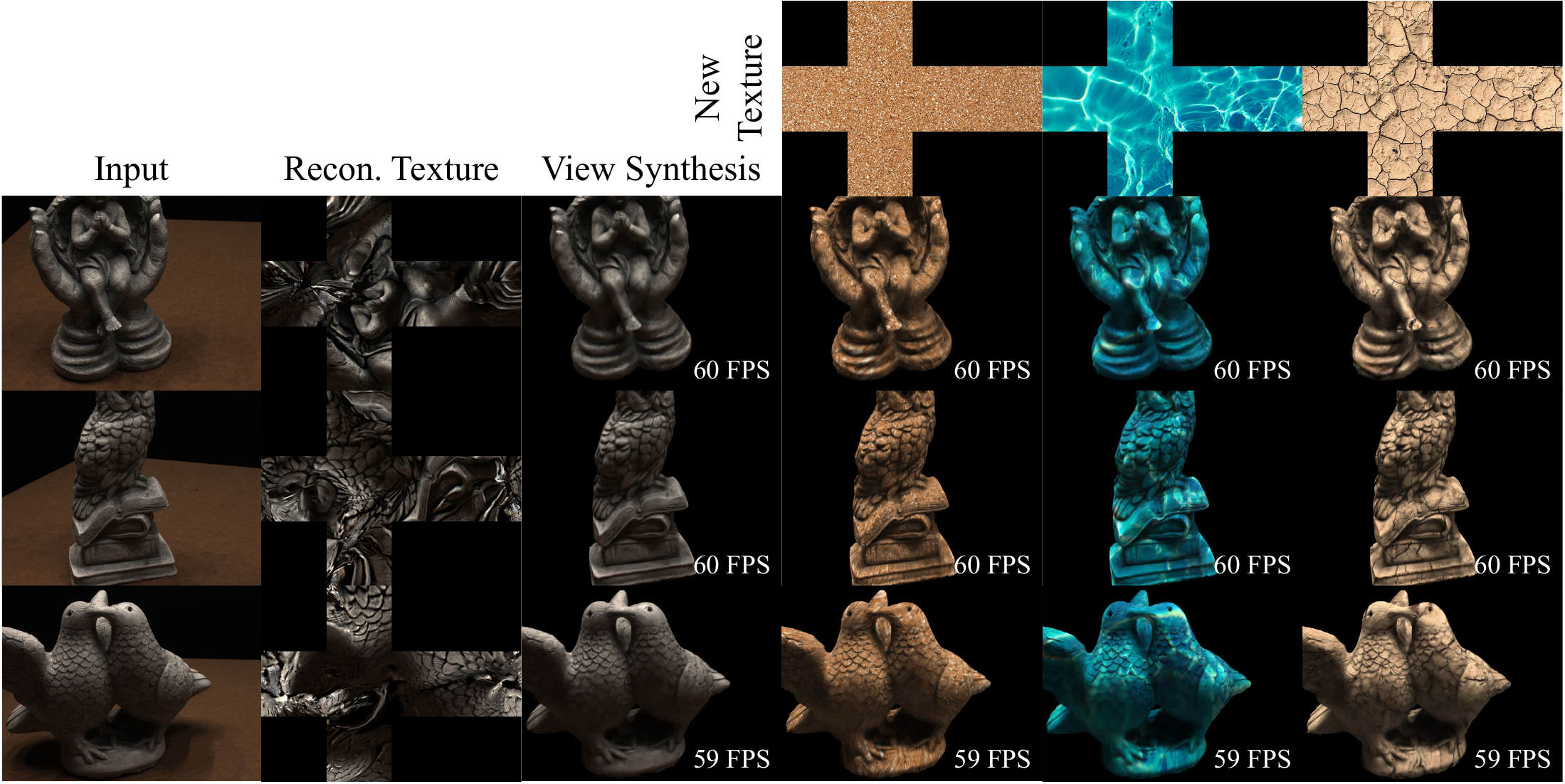}
  
    \vspace{-0.3cm}
    \caption{Texture swapping with our method. We propose to disentangle the appearance from the geometry for 3D-GS, thereby facilitating real-time appearance editing such as texture swapping. The rendering speed is shown in each result.}
  \vspace{-0.6cm}
  \label{fig:teaser}
\end{figure}

Reconstruction, editing, and real-time rendering of photo-realistic scenes are fundamental problems in computer vision and graphics, with diverse applications such as film production, computer games, and virtual/augmented reality. 
Polygonal meshes have served as the standard 3D representation within traditional rendering pipelines, owing to their rendering speed and editing flexibility (with texture mapping). 

Due to the laborious process of manual mesh-based scene modeling, 3D Gaussian Splatting~\cite{3dgs} (3D-GS) has gained considerable attention for its capability of faithfully reconstructing complex scenes from multi-view images and real-time rendering.
3D-GS represents the scene as a set of 3D anisotropic Gaussians equipped with per-Gaussian color attributes and supports real-time rendering by splatting these Gaussians onto the image plane.
However, this representation couples the appearance and geometry of the scene within the unordered and irregular 3D Gaussians, which hinders the flexibility of appearance editing for 3D-GS compared to meshes, where appearance can be easily parameterized into texture maps.
Although considerable efforts have been made to edit 3D Gaussian-based scenes~\cite{ye2023gaussian,gaussianeditor1,gaussianeditor2,physgaussian,kirillov2023segment}, the manipulation of appearance remains inconvenient since these works still follow the entangled representation of 3D-GS.

In this paper, we propose a novel method, named Texture-GS, which aims to explicitly disentangle the geometry and texture for 3D-GS, thereby significantly improving the flexibility of appearance editing for 3D scenes.
Texture-GS follows 3D-GS in modeling the geometry as a set of anisotropic 3D Gaussians, but crucially, it represents the view independent appearance as a 2D texture map.
Leveraging this disentangled representation, Texture-GS retains the powerful capabilities of 3D-GS for faithful reconstruction and real-time rendering, while also facilitating various appearance editing applications, such as texture swapping shown in \cref{fig:teaser}.

The key challenge in implementing our Texture-GS lies in establishing a connection between the geometry (3D Gaussians) and appearance (2D texture map).
NeuTex~\cite{neutex} has proposed a texture mapping MLP (Multi-Layer Perceptron) that regresses 2D UV coordinates for every 3D point to represent the radiance of NeRF (Neural Radiance Field)~\cite{nerf} in a 2D texture space.
However, evaluating an MLP for each ray-Gaussian intersection is unsuitable for our Texture-GS, as it would be prohibitively
expensive for real-time rendering, which is a key advantage of 3D-GS over NeRF-based methods.
To maintain the ability of real-time rendering, one straightforward solution is to employ a texture mapping MLP to pre-compute UV coordinates for each Gaussian based on its center location and then query the \emph{per-Gaussian} color attributes from the texture map  before rendering.
Given the fact that each 3D Gaussian often covers more than one pixel in practice, this straightforward solution would result in all pixels covered by a single Gaussian being mapped to the same UV location, leading to discontinuities in the texture space.
To address this issue, we propose a novel \emph{texture mapping module}.
It incorporates an MLP for mapping Gaussian centers into the texture UV space before rendering, along with a Taylor expansion at the Gaussian centers, which serves an approximation of the MLP and enables efficient mapping of the ray-Gaussian intersections to UV coordinates during rendering.
Our texture mapping module not only promotes a smooth texture map, as the Taylor expansion guarantees the local continuity for the UV coordinates of pixels within a projected Gaussian, but also preserves rendering efficiency, as the inference of UV coordinates merely involves a small matrix product.
 
To evaluate the effectiveness of our Texture-GS, we conduct extensive quantitative and qualitative experiments on the DTU dataset~\cite{dtu}.
The results demonstrate that Texture-GS recovers a smooth high-quality 2D texture map from multi-view images, while also facilitates various editing applications such as global texture swapping and fine-grained texture editing. Besides, our method achieves an average rendering speed of 58 FPS on a single RTX 2080 Ti GPU, demonstrating its real-time rendering capabilities. 
Our contributions are summarized below.
\begin{itemize}
    \item To the best of our knowledge, we are the first to disentangle the geometry and texture for 3D-GS, thereby enabling various editing applications. 
    \item We propose a novel texture mapping module to map ray-Gaussian intersections into a continuous 2D texture space while maintaining efficiency.
    \item Experiments validate the effectiveness of our method for novel view synthesis, global texture swapping, and local appearance editing, with real-time rendering speed on consumer-level devices.
\end{itemize}

\section{Related Work}

\noindent\textbf{Neural UV Mapping.}
UV mapping plays an essential role in the traditional rendering pipeline, aiming at computing a bijective mapping between the 3D surface and a suitable parametric domain. UV mapping is usually accompanied with 3D shapes and jointly modeled by artists, necessitating considerable labor costs. In recent years, NeRF~\cite{nerf} has gained increasing attention for its superior view synthesis quality, inspiring many follow-up works~\cite{neutex,ngf,IsometricSurface} to reconstruct 3D geometry with a volumetric density field while concurrently learning UV mapping based on neural networks. NeuTex~\cite{neutex} is the first work to recover a meaningful surface-aware UV mapping function from multi-view images. Provided with any 3D shading point during the NeRF's ray marching process, NeuTex obtains its radiance by sampling the reconstructed texture at its UV location, which is output by a texture mapping MLP. To ensure the bijective property of UV mapping, NeuTex adopts a cycle consistency loss to regularize the network. The follow-up Neural Gauge Field~\cite{ngf} draws inspiration from the principle of information conservation during gauge transformation~\cite{gaugetheory} and proposes an information regularization term to maximize the mutual information. Nuvo~\cite{nuvo} extends NeuTex with multiple charts and a chart assignment network for general scenes. Apart from the above methods, some approaches focus on specific object categories such as human faces~\cite{xu2023bi,ma2022neural,auvnet}, documents~\cite{IsometricSurface} and human bodies~\cite{chen2023uv}. However, NeRF-based methods adopt ray marching for rendering, which evaluates a large texture MLP with an additional UV mapping network at hundreds of sample shading points along the ray for each pixel to compute the final color. This process is excessively slow for interactive visualization and real-time applications.  

\vspace{3mm}
\noindent\textbf{3D Gaussian Editing.}
3D Gaussian Splatting~\cite{3dgs} (3D-GS) has emerged as an alternative 3D representation to NeRF~\cite{nerf}, achieving real-time rendering via splatting 3D Gaussians instead of ray marching. It has received increasing attention for its explicit representation and promising reconstruction quality, which is more suitable for scene editing.  Leveraging its explicit, point cloud-like formulation, Point’n Move~\cite{huang2023point}, Gaussian Grouping~\cite{ye2023gaussian}, SA-GS~\cite{hu2024semantic} and Feature 3DGS~\cite{zhou2023feature} combine semantic segmentation methods such as SAM~\cite{kirillov2023segment} with 3D GS representation to obtain the mask of the target and explicitly manipulate the selected object in the scene in real time. SC-GS~\cite{huang2023sc} and Cogs~\cite{yu2023cogs} introduce novel frameworks for manipulating and editing dynamic content in 4D space. With the advancement of text-to-image models~\cite{rombach2022high}, some works~\cite{gaussianeditor1,gaussianeditor2} achieve swift and controllable 3D scene editing in accordance with text instructions, incorporating a 2D diffusion model to fine-tune 3D-GS representations. PhysGaussian~\cite{physgaussian} explores the physical properties of 3D Gaussians, employing a custom Material Point Method for physical simulation. Despite yielding promising results, these methods capture the appearance in per-Gaussian color attributes and regard 3D Gaussians as isolated shading elements, neglecting the global appearance. The entanglement of color and density attributes hinders editing flexibility and precludes editing applications such as texture swapping. 
\section{Preliminaries}

3D-GS~\cite{3dgs} represents the scene as a set of 3D Gaussians $\mathcal{G} = \{G_i(x)\}_{i=1}^{N}$, where $N$ denotes the number of Gaussians. 
Each Gaussian is defined by its center position $\mu_i \in \mathbb{R}^3$ and covariance matrix $\Sigma_i \in \mathbb{R}^{3\times 3}$, expressed as:
\begin{equation}
    G_i(x) = \exp\left({-\frac{1}{2}(x-\mu)^T\Sigma_i^{-1}(x-\mu)}\right).
\end{equation}
% Here $\Sigma_i$ is factorized into a scaling matrix $S_i\in \mathbb{R}^{3\times 3}$ and a rotation matrix $R_i \in \mathbb{R}^{3\times 3}$ as $\Sigma_i = R_iS_iS_i^TR_i^T$. $S_i=\text{diag}(s_i)$ is provided by a scaling vector $s_i\in \mathbb{R}^3$ and $R_i$ is represented by a unit quaternion $q_i\in \textbf{SO(3)}$. 
% 
The appearance of the scene is represented in the per-Gaussian attributes, \ie an opacity value $o_i \in \mathbb{R}$ to adjust the influence weight and an RGB color $c_i \in \mathbb{R}^3$ given by sphere harmonic (SH) coefficients.
To render the scene, 
% Instead of sampling shading points, 
3D-GS splats 3D Gaussians onto the image plane via EWA splatting~\cite{zwicker2002ewa} to form 2D Gaussians $G_i'$, whose covariance matrix $\Sigma_i'\in \mathbb{R}^{2\times 2}$ is defined as $\Sigma_i' = JW\Sigma_iW^TJ^T $.
Here $W\in \mathbb{R}^{3\times 3}$ denotes the viewing transformation matrix and $J\in \mathbb{R}^{2\times 3}$ is the Jacobian matrix of the affine approximation of the perspective projection.
% 
% For efficient rendering, each Gaussian is equipped with an opacity value $o_i \in \mathbb{R}$ to adjust the influence weight and an RGB color $c_i \in \mathbb{R}^3$ given by sphere harmonic (SH) coefficients. 
% 
The final color $C_p$ at pixel $p$ is given by a set of ordered Gaussians with cumulative volumetric rendering:
\begin{equation}
    C_p = \sum_{j\in \mathcal{N}_p} c_j \alpha_j \prod_{k=1}^{j-1} (1-\alpha_k),
    \label{eq:alpha-blending}
\end{equation}
where $\mathcal{N}_p$ is the number of projected Gaussians within a title, 
%%%YKL Why title here? Do you mean pixel?
and $\alpha_j = o_j G_j'(p)$ is the transparency value.
% 
% Here the transparency value $\alpha_j$ is defined as $\alpha_j = o_j G_j'(p)$. 
% 
Notably, the geometry (the positions and scales of 3D Gaussians) and appearance (per-Gaussian colors) are entangled in the 3D-GS representation, thereby complicating downstream appearance editing applications such as texture swapping.
\section{Method}

\subsection{Overview}

Given the multi-view images of a real-world scene, the objective of our Texture-GS is to reconstruct the geometry using a set of 3D Gaussians and the appearance through our proposed \emph{texture mapping module}, in a disentangled manner, enabling flexible texture editing applications.
Our texture mapping module consists of a UV mapping MLP that projects 3D points into 2D UV space, and a learnable 2D texture representing the appearance of the 3D scene.
It is highly under-constrained to jointly learn the UV mapping MLP and the texture values.
Therefore, we first learn the UV mapping MLP using a cycle consistent constraint on the bijective mapping between the 3D space and the UV texture domain (in \cref{sec:uv_map}) and then learn the 2D texture values using a photometric loss and Taylor-expansion-based approximation of the MLP for efficiency (in \cref{sec:tex_gs}).

\subsection{Learning UV Mapping MLP}
\label{sec:uv_map}

The UV mapping MLP, denoted as $\phi: \mathbb{R}^3 \rightarrow \mathbb{R}^2$, aims to regress 2D UV coordinates $u\in \mathbb{R}^2$ from the 3D point $x \in \mathbb{R}^3$ that lies either on or close to the underlying surface of the scene.
It is non-trivial to learn the highly under-constrained mapping $\phi$ without paired training data.
Intuitively, we observe that if its inverse 2D-to-3D function, $\phi^{-1}: \mathbb{R}^3 \rightarrow \mathbb{R}^2$, can warp the UV space to uniformly cover the underlying surface, then the forward mapping $\phi$ would serve as a reasonable UV mapping.

Based on this observation, we should first estimate the underlying surface of the object from multi-view images, before the UV mapping learning.
To this end, we train a vanilla 3D-GS and leverage the alpha-blending aggregation in \cref{eq:alpha-blending} to obtain the depth value $D_p$ for pixel $p$:
\begin{equation}
    D_p = \sum_{j\in \mathcal{N}_p} d_j \alpha_j \prod_{k=1}^{j-1} (1-\alpha_k),
    \label{eq:alpha-blending-depth}
\end{equation}
where $d_j = (R \mu_j + t)_z$ denotes the depth value of the center $\mu_j$ in the camera coordinate system.
Given the assumption that the editing target has an opaque and smooth surface,
the back-projected 3D points $\{x_i\}_{i=1}^{N_d}$ from these depth maps would lie either on or close to the underlying surface.

With these 3D points approximating the underlying surface, we can establish the following three constraints to learn the UV mapping $\phi$.
(1) We introduce a cycle consistency loss $\mathcal{L}_\text{cycle}^\text{3d}$ in the 3D space to ensure the mutually inverse property of $\phi$ and $\phi^{-1}$:
\begin{equation}
    \mathcal{L}_\text{cycle}^\text{3d} = \frac{1}{N_d}\sum_{i=1}^{N_d}||x_i-\phi^{-1}\circ\phi(x_i)||.
    \label{eq:cycle_loss}
\end{equation}
(2) To encourage $\phi^{-1}$ to wrap the underlying surface with a uniform 2D UV plane, we randomly sample a set of evenly distributed UV coordinates $\mathcal{U} = \{u_i\}_{i=1}^{N_u}$, and then minimize the symmetric Chamfer distance between the 3D points mapped with $\phi^{-1}$ from $\mathcal{U}$ and the pseudo ground truth point cloud $\mathcal{P}=\{p_i\}_{i=1}^{N_p}$ extracted from the trained 3D-GS via the farthest point sampling on the centers of the 3D Gaussians:
\begin{equation}
    \mathcal{L}_\text{CD} = \frac{1}{N_u}\sum_{i=1}^{N_u}\min_{p_j\in \mathcal{P}}||\phi^{-1}(u_i)-p_j|| + \frac{1}{N_p} \sum_{j=1}^{N_p}\min_{u_i\in \mathcal{U}}||\phi^{-1}(u_i)-p_j||.
    \label{eq:cd_loss}
\end{equation}
(3) We also leverage the sampled UVs $\mathcal{U}$ to regularize the bijectivity via the cycle consistency loss $\mathcal{L}_\text{cycle}^\text{2d}$ in the 2D space:
\begin{equation}
    \mathcal{L}_\text{cycle}^\text{2d} = \frac{1}{N_u}\sum_{i=1}^{N_u}||u_i-\phi\circ\phi^{-1}(u_i)||.
    \label{eq:cycle_loss2}
\end{equation}
These three constraints are combined together to learn the UV mapping MLP:
\begin{equation}
    \mathcal{L}_\text{UV} = \mathcal{L}_\text{cycle}^\text{3d} + \mathcal{L}_\text{CD} + \mathcal{L}_\text{cycle}^\text{2d}.
\end{equation}
Although NeuTex~\cite{neutex} also utilizes the cycle consistency constraint for learning the UV mapping, it computes the loss term with each sampled 3D shading point on the rays. In contrast, our method only applies constraints to 3D points near the underlying surface, which results in higher efficiency for training.

\begin{figure}[tb]
    \centering
    \includegraphics[width=\linewidth]{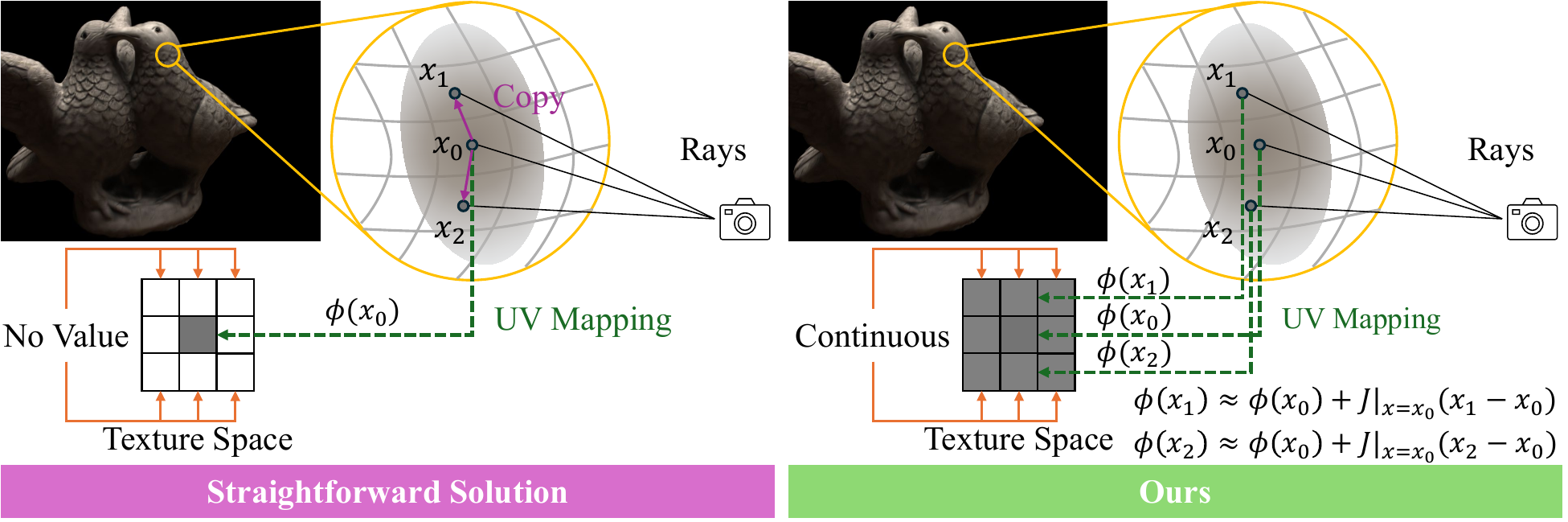}
    \vspace{-0.6cm}
    \caption{Comparison with the straightforward solution. The straightforward solution fails to generate a reasonable and continuous texture, while our method can reconstruct a high-quality texture by considering the intersection of 3D Gaussians and each ray.}
    \vspace{-0.6cm}
    \label{fig:problem}
\end{figure}

\subsection{Learning Texture Value}
\label{sec:tex_gs}
Having the UV mapping MLP $\phi$, our goal is to reconstruct the high-quality texture from multi-view images through differentiable rendering, enabling flexible scene appearance editing, \eg texture swapping, and real-time rendering.
It is not practical to
predict UV coordinates for each 3D point using $\phi$, akin to the NeuTex~\cite{neutex}, as evaluating an MLP for each ray-Gaussian intersection consumes too much computational cost for real-time rendering.
To this end, a straightforward solution is to apply the UV mapping MLP $\phi$ to each Gaussian's center to pre-compute UV coordinates, then query the \emph{per-Gaussian} color attributes from the learnable texture before rendering, and finally optimize the texture value through differentiable Gaussian splatting.
However, since each 3D Gaussian often covers multiple pixels in practice, this solution would map all pixels covered by a single Gaussian to the same UV location (\cref{fig:problem}, left), resulting in degenerated textures. 
This would significantly hinder appearance editing.

To address this issue, we propose to allow pixels covered by a single Gaussian to have different UV coordinates (\cref{fig:problem}, right), which encourages a smooth texture for the convenience of appearance editing.
We rewrite the rendering equation in \cref{eq:alpha-blending} as:
\begin{equation}
    C_p = \sum_{j\in \mathcal{N}_p} \mathcal{C}(G_j, r_p) \alpha_j \prod_{k=1}^{j-1} (1-\alpha_k).
\end{equation}
Here, we replace original per-Gaussian color attributes with a color function $\mathcal{C}$, which takes both the 3D Gaussian $G_j$ and the ray $r_p = o + t\dot d_p$ along the pixel $p$ as input, where $o\in \mathbb{R}^3$ is the viewpoint location and $d_p \in \mathbb{R}^3$ is a unit view direction vector.

\paragraph{Ray-Gaussian intersection.}
To establish the color function $\mathcal{C}$, we initially compute the intersection $I(G_j, r_p)$ between a ray $r_p$ and a 3D Gaussian $G_j$.
However, the vanilla 3D Gaussians denote a volume density function without solid surfaces, hindering the derivation of ray-Gaussian intersection.
Fortunately, in most practical cases, the Gaussian should be flat and aligned closely with the opaque surface, as mentioned in previous work~\cite{guedon2023sugar,jiang2023gaussianshader}.
To achieve this, we introduce a zero-one regularization term to the opacity value $o_i$
\begin{equation}
    \mathcal{L}_{01} = \frac{1}{N} \sum_{i=1}^{N} (\ln(o_i) + \ln(1-o_i)),
\end{equation}
and prune those Gaussians with an opacity that is less than 0.5.
Subsequently, we flatten the 3D Gaussian along its normal vector $n_j$, which is defined as the eigenvector $v_j$ associated with the smallest eigenvalue of the covariance matrix $\Sigma_j$.
To ensure the normal $n_j$ points outwards from the surface, we selectively reverse $v_j$ based on the view direction $\mu_j - o$:
\begin{equation}
\begin{aligned}
    n_j = \left\{\begin{matrix}
        v_j,  & \text{if} \quad v_j \cdot (\mu_j - o) < 0, \\
        -v_j, & \text{otherwise}.
    \end{matrix}\right.
    \end{aligned}
\end{equation}
Hence, the intersection between the flattened Gaussian and the ray is given by:
\begin{equation}
    I(G_j, r_p) = o + \frac{(\mu_j - o)\cdot n_j}{d_p \cdot n_j} d_p.
\end{equation}
As this derivation relies heavily on the normal vector $n_j$ of 3D Gaussian, we additionally introduce a normal loss to supervise the rendered normal map $\overline{N}$ with the pseudo ground truth $\overline{N}_\text{gt}$ derived from SfM~\cite{schoenberger2016sfm} result under the local planarity assumption:
\begin{equation}
    \mathcal{L}_\text{norm} = \frac{1}{HW} ||\overline{N}-\overline{N}_\text{gt}||_2^2.
\end{equation}
To suppress high-frequency noise in $\overline{N}_\text{gt}$, we also apply a spatial smoothness regularization term:
\begin{equation}
    \mathcal{L}_\text{sm} = \frac{1}{HW} \sum_{p} \sum_{q\in \mathcal{N}(p)} \exp(-\gamma ||C_\text{gt}(p)-C_\text{gt}(q)||_1) ||\overline{N}(p)-\overline{N}(q)||_1.
\end{equation}
Besides, to avoid the intersection far away from the center of 3D Gaussian, we clamp the intersection with a radius defined by the eigenvalue of the matrix $\Sigma_i$.

\paragraph{Efficient UV mapping.} Due to the high computational cost of evaluating the MLP $\phi$ for each ray-Gaussian intersection, as shown in the right part of \cref{fig:problem}, we propose to approximate the UV coordinates $\phi(I(G_j, r_p))$ utilizing the first two terms of the Taylor expansion (i.e., first-order approximation) of $\phi$ at each Gaussian center $\mu_j$, written as
\begin{equation}
    \tilde{\phi}(I(G_j, r_p)) = \phi(\mu_j) + J|_{x=\mu_{j}}(I(G_j, r_p) - \mu_j),
    \label{Taylor}
\end{equation}
where $J|_{x=\mu_{j}}$ denotes the Jacobian matrix of $\phi$ at the point $\mu_j$.
The Jacobian matrix can be efficiently obtained with automatic differentiation and pre-computed for each Gaussian center before rendering.
Thus, \cref{Taylor} solely involves a small matrix product, significantly reducing the computational cost during rendering, and thereby enabling real-time rendering for downstream applications. 

\paragraph{Color function.} Leveraging the UV coordinates for each intersection, we can directly query the texture $\mathcal{T}$ to establish the color function $\mathcal{C}$.
Notably, view-dependent appearance is common in real-world scenes, typically caused by non-Lambertian materials.
3D-GS~\cite{3dgs} addresses this issue by representing the per-Gaussian color as an ordered set of spherical harmonic (SH) coefficients, where we interpret the first SH coefficient as the diffuse term and others as the view-dependent appearance.
However, representing the textures with SH coefficients consumes high memory and computational cost, as the resolution of textures is usually much larger than the number of 3D Gaussians.
Therefore, we propose to represent the diffuse color within the texture $\mathcal{T}$ and utilize per-Gaussian SH coefficients $c^{\text{SH}}_j$ for the residual view-dependent appearance, written as:
\begin{equation}
    \mathcal{C}(G_j, r_p) = h( \tilde{\phi}(I(G_j, r_p)), \mathcal{T}) + c^{\text{SH}}_j,
    \label{eq:hybrid_func}
\end{equation}
where $h(\cdot, \mathcal{T})$ denotes the sampling on the texture $\mathcal{T}$ with UV coordinates using the bilinear interpolation.

The final loss for supervision is defined as:
\begin{align}
    \mathcal{L}           & = \mathcal{L}_\text{GS} + \lambda (\mathcal{L}_\text{1}^\text{noSH} + \lambda_\text{ssim} \mathcal{L}_\text{ssim}^\text{noSH}), \\
    \text{where}\;\;\;\; \mathcal{L}_\text{GS} & = \mathcal{L}_1 +  \mathcal{L}_\text{mask} + \lambda_\text{ssim} \mathcal{L}_\text{ssim} + \lambda_\text{01}  \mathcal{L}_{01} + \lambda_\text{n} (\mathcal{L}_\text{norm} + \mathcal{L}_\text{sm}).
\end{align}
$\mathcal{L}_1$ and $\mathcal{L}_\text{ssim}$ are the L1 and D-SSIM loss terms to regularize the visual quality of rendered images, as in 3D-GS~\cite{3dgs}. $\mathcal{L}_\text{mask}$ is the mask loss when viewpoints do not entirely cover the object, as in NeuTex~\cite{neutex}. $\mathcal{L}_\text{1}^\text{noSH}$ and $\mathcal{L}_\text{ssim}^\text{noSH}$ are the L1 and D-SSIM loss terms for rendering without per-Gaussian SH coefficients to encourage most (especially view-independent) appearance information to be stored in the texture.
Please refer to the supplementary material for more details.

\section{Experiments}

We conduct experiments on real scenes from DTU dataset~\cite{dtu} to validate the effectiveness of our model, wherein each scene contains 49 or 64 input images captured from different views. We utilize the foreground masks provided by IDR~\cite{idr} to segment the object. Following NeuTex~\cite{neutex}, we sample 5 single-object scenes for quantitative experiments, withhold 4 views as test data, and train our method with the remaining images for novel view synthesis. For evaluation, we report three commonly adopted performance metrics, including Peak Signal-to-Noise Ratio (PSNR), L1 metric and Learned Perceptual Image Patch Similarity (LPIPS). All evaluation experiments are conducted on a single RTX 2080 Ti GPU. More details can be seen in the supplementary material.

\subsection{Novel View Synthesis}

\begin{figure}[tb]
  \centering
  \includegraphics[width=\linewidth]{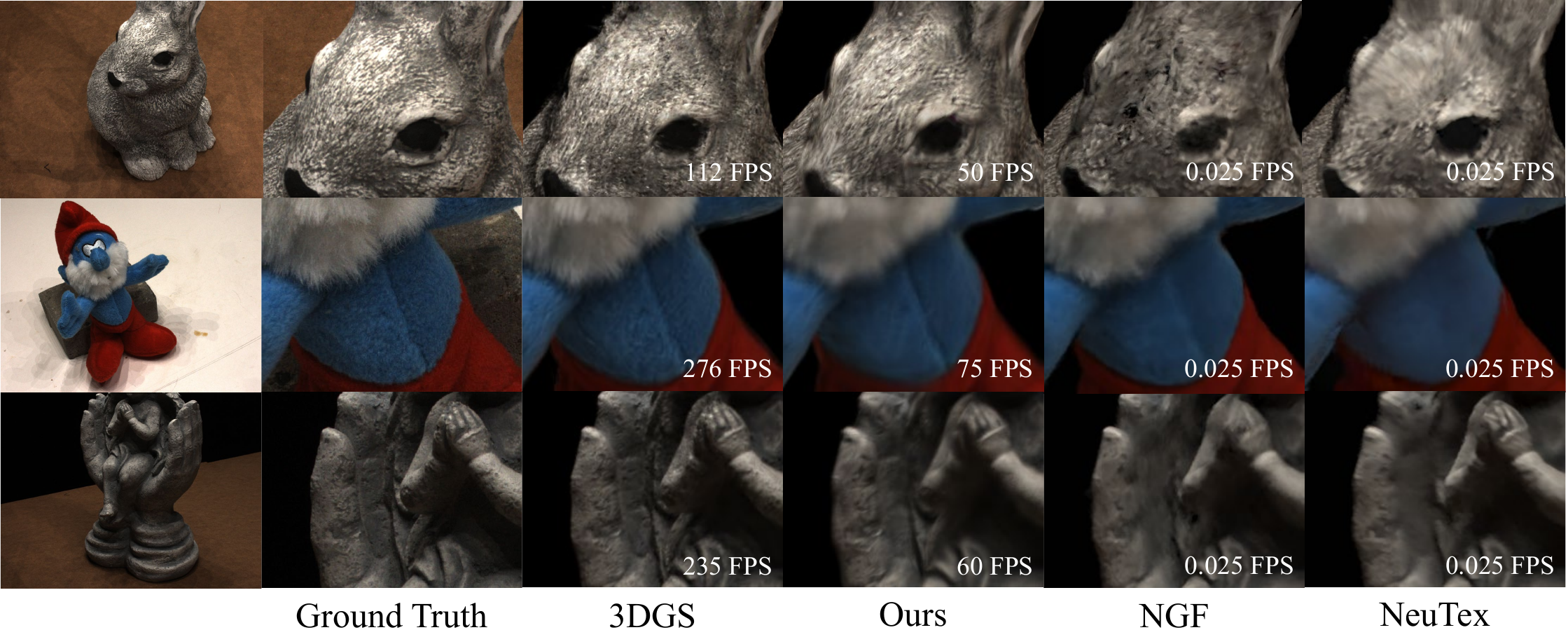}
  \vspace{-0.6cm}
  \caption{Visual comparison with previous state-of-the-art editing methods}
  \vspace{-0.3cm}
  \label{fig:nvs_sota}
\end{figure}

We present comprehensive comparisons of our method with previous state-of-the-art (SOTA) editing methods that decouple the geometry and appearance, including NeRF-based NeuTex~\cite{neutex} and Neural Gauge Fields~\cite{ngf} (NGF) on novel view synthesis. Although NeuMesh~\cite{neumesh} and Seal-3D~\cite{seal3d} also support appearance editing applications, they require an extra finetuning stage (ranging from seconds to hours) to inject the edited appearance into the 3D representation, which is inconsistent with real-time editing objectives. Naturally, our method is more constrained than the vanilla 3D-GS~\cite{3dgs}, hence we also compare our method with 3D-GS to investigate the loss of rendering quality and computational efficiency. 

\cref{fig:nvs_sota} illustrates the visual comparison on zoom-in crops of test images. Although the rendering quality is slightly inferior to 3D-GS, our method can still synthesize photo-realistic rendered images. Compared with NeRF-based approaches, our method achieves superior performance, especially in regions observable from merely a few views of training data, such as the rabbit's head and the inner side of the palm. Thanks to the explicit geometry of 3D Gaussians, Texture-GS can generate higher-quality view synthesis results, while NeRF requires more views to learn the implicit representation. Most importantly, our method achieves real-time rendering speed, enabling instant preview of the editing results to facilitate interactive editing applications. 

\begin{table}[tb]
  \caption{Comparison of novel view synthesis results on the DTU dataset.}
  
    \vspace{-0.6cm}
  % \label{tab:nvs_sota}
  \centering
    \setlength{\tabcolsep}{2pt}
    \begin{subtable}[t]{0.49\linewidth}
        \caption{Comparison with the SOTAs}
        
        \vspace{-0.3cm}
        \label{tab:nvs_sota}
        \centering
        \resizebox{0.9\linewidth}{!}{
            \begin{tabular}{c|cccc}
                \toprule
                \multirow{2}{*}{Method} & \multicolumn{4}{c}{DTU} \\
                & PSNR↑ & L1↓ & LPIPS↓ & FPS\\
                \midrule
                NeuTex & 30.39 & 0.0158 & 0.1613 & 0.025\\
                NGF & 29.44 & 0.0166 & 0.1506 & 0.025 \\
                3DGS & 30.99 & 0.0121 & 0.1079 & 198\\
                Ours & 30.03 & 0.0135 & 0.1440 & 58\\
              \bottomrule
            \end{tabular}
        }
    \end{subtable}
    \begin{subtable}[t]{0.49\linewidth}
        \caption{Different number of 3D Gaussians}
        
        \vspace{-0.3cm}
        \centering
        \label{tab:num_sota}
        \resizebox{0.9\linewidth}{!}{
            \begin{tabular}{c|cccc}
                \toprule
                \multirow{2}{*}{\#Gauss} & \multicolumn{4}{c}{DTU} \\
                & PSNR↑ & L1↓ & LPIPS↓ & FPS\\
                \midrule
                100\% & 30.03 & 0.0135 & 0.1440 & 58\\
                50\% & 29.57 & 0.0142 & 0.1555 & 69 \\
                20\% & 28.75 & 0.0155 & 0.1705 & 82\\
                5\% & 27.86 & 0.0172 & 0.1841 & 104\\
              \bottomrule
          \end{tabular}
        }
     \end{subtable}
    \vspace{-0.5cm}
\end{table}

We present the quantitative comparison of the average metrics in \cref{tab:nvs_sota}. Our method is 2000× faster than NeRF-based methods, maintaining the rendering quality that is on par and sometimes superior on the same hardware. The average training time of NeuTex~\cite{neutex} is 30 hours per scene, while ours amounts to 90 minutes. 

\begin{figure}[tb]
% \vspace{-0.4cm}
  \centering
  \includegraphics[width=\linewidth]{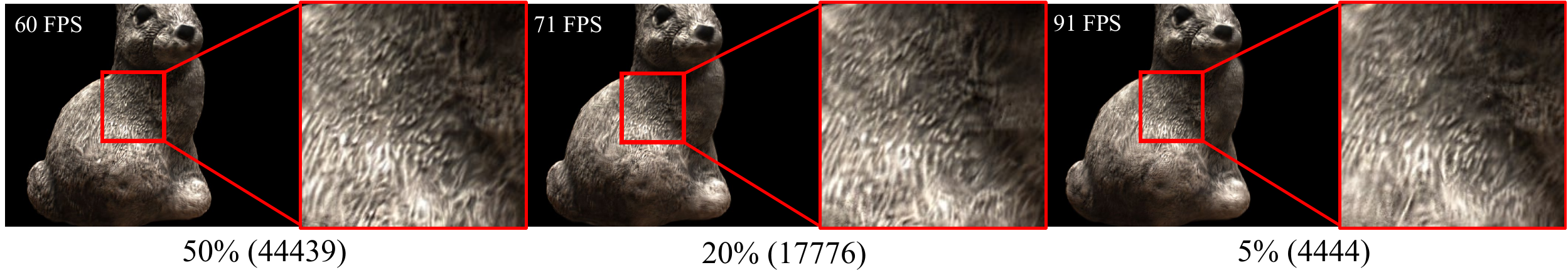}
  
\vspace{-0.3cm}
  \caption{Visual comparison of our method with different numbers of 3D Gaussians}
  
\vspace{-0.3cm}
  \label{fig:num_sota}
\end{figure}

Texture-GS allows pixels covered by a single Gaussian to possess different colors, thereby enhancing the representational power for each Gaussian. To further accelerate the rendering speed, we simplify the geometry by pruning 50\%, 80\% and 95\% 3D Gaussians according to their opacity values. As illustrated in \cref{tab:num_sota}, reducing the number of Gaussians inevitably leads to a minor performance decline, while greatly reducing the computational costs. \cref{fig:num_sota} shows that Texture-GS can synthesize high-quality images for objects with rich texture even with 4.4K 3D Gaussians representing the geometry, which demonstrate the potential extensibility to a wide range of computing platforms.

\subsection{Texture Editing}

\begin{figure}[tb]
  \centering
    \begin{subfigure} {\linewidth}
        \includegraphics[width=\linewidth]{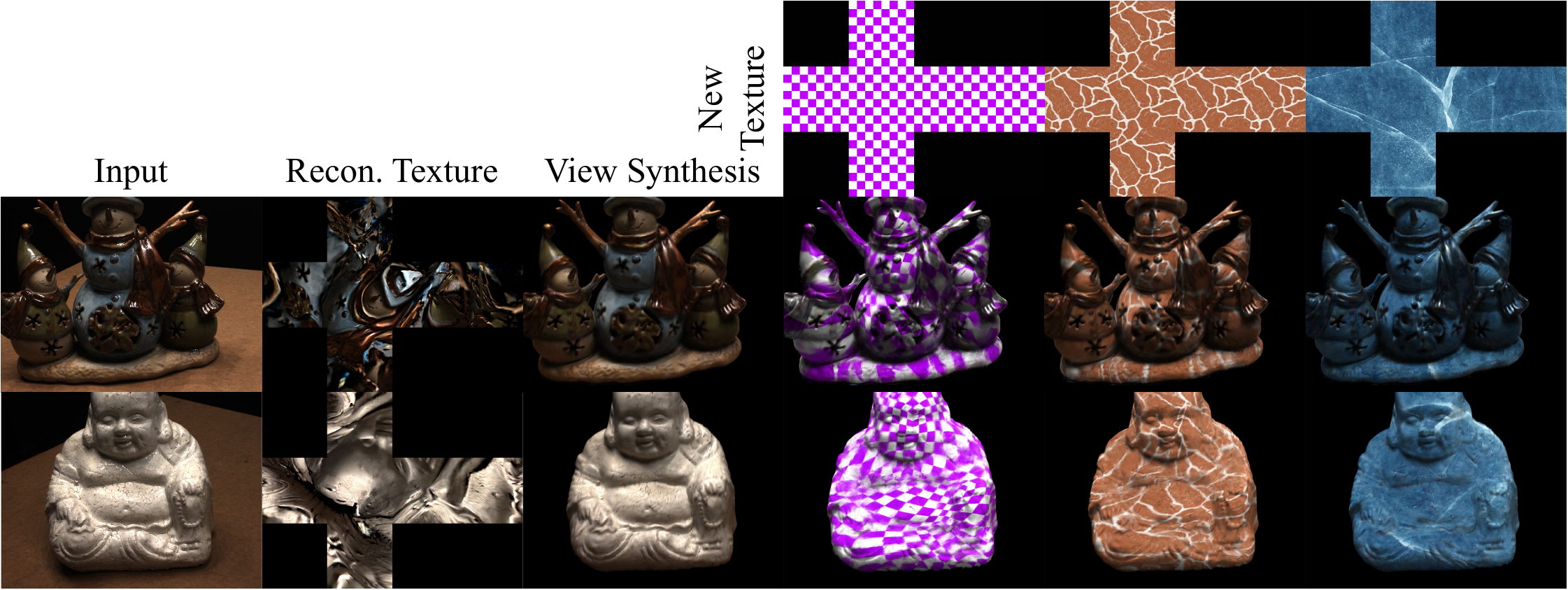}
        \caption{Under various textures}
        \label{fig:tex_swapping}
    \end{subfigure}
    \begin{subfigure} {\linewidth}
        \centering
        \includegraphics[width=\linewidth]{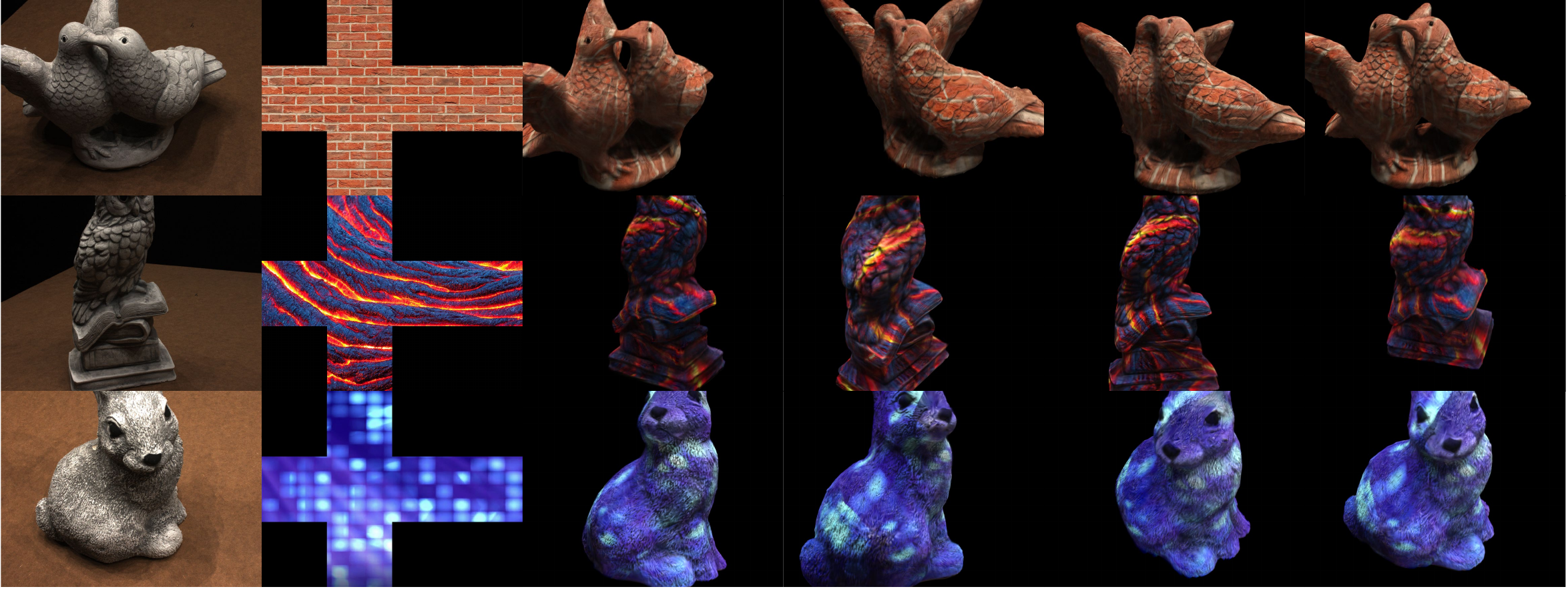}
        \caption{Under different views}
        \label{fig:tex_swapping2}
    \end{subfigure}
    
\vspace{-0.3cm}
  \caption{Visualization of texture swapping results of our method}
\vspace{-0.6cm}
\end{figure}

\begin{figure}[tb]
% \vspace{-0.3cm}
  \centering
  \includegraphics[width=\linewidth]{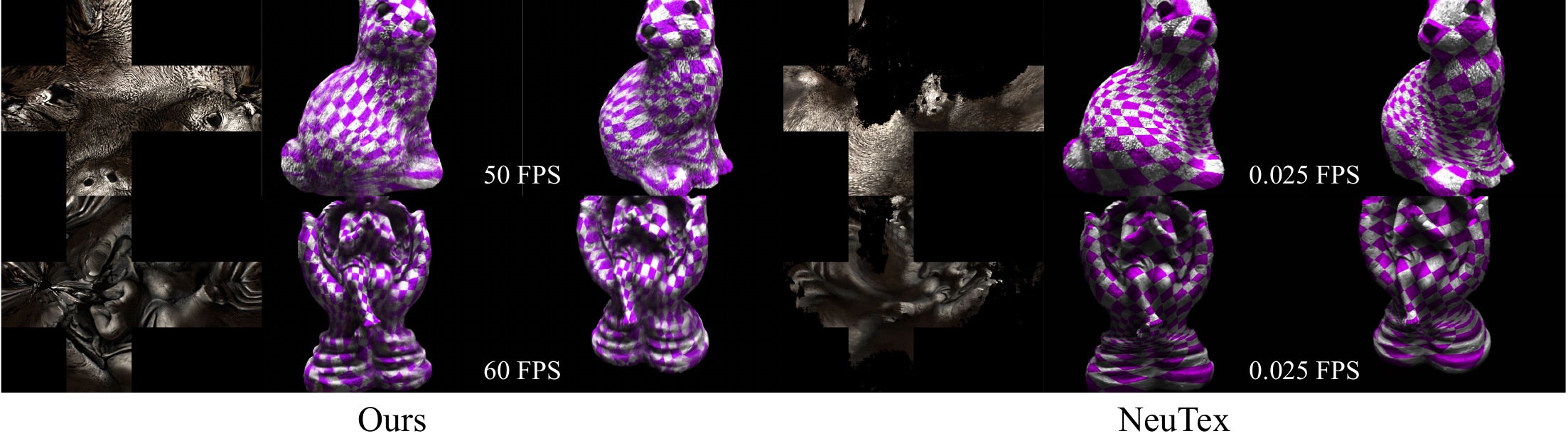}
  
\vspace{-0.3cm}
  \caption{Visual comparison of our method with NeuTex~\cite{neutex} on texture swapping.}
  
\vspace{-0.3cm}
  \label{fig:tex_swap_sota}
\end{figure}

\noindent\textbf{Texture Swapping.} Leveraging the decoupled representation, we can support real-time appearance editing such as changing the global texture of reconstructed objects. As shown in \cref{fig:tex_swapping}, our method can render photo-realistic and sharp images under various textures. Even for the porcelain with complex geometry structure, our method can reconstruct a high-quality texture from multi-view images. \cref{fig:tex_swapping2} shows the novel view synthesis results of texture swapping, highlighting our method's capability of maintaining the consistency of the geometry and appearance under different views. We also present a visual comparison with NeuTex~\cite{neutex} in Fig.~\ref{fig:tex_swap_sota}. Our method learns a more uniform texture space while achieving real-time rendering speed. The results emphasize the practical utility of Texture-GS in applications requiring real-time 3D scene manipulation and visualization.

\begin{figure}[tb]
  \centering
  
% \vspace{-0.5cm}
  \includegraphics[width=\linewidth]{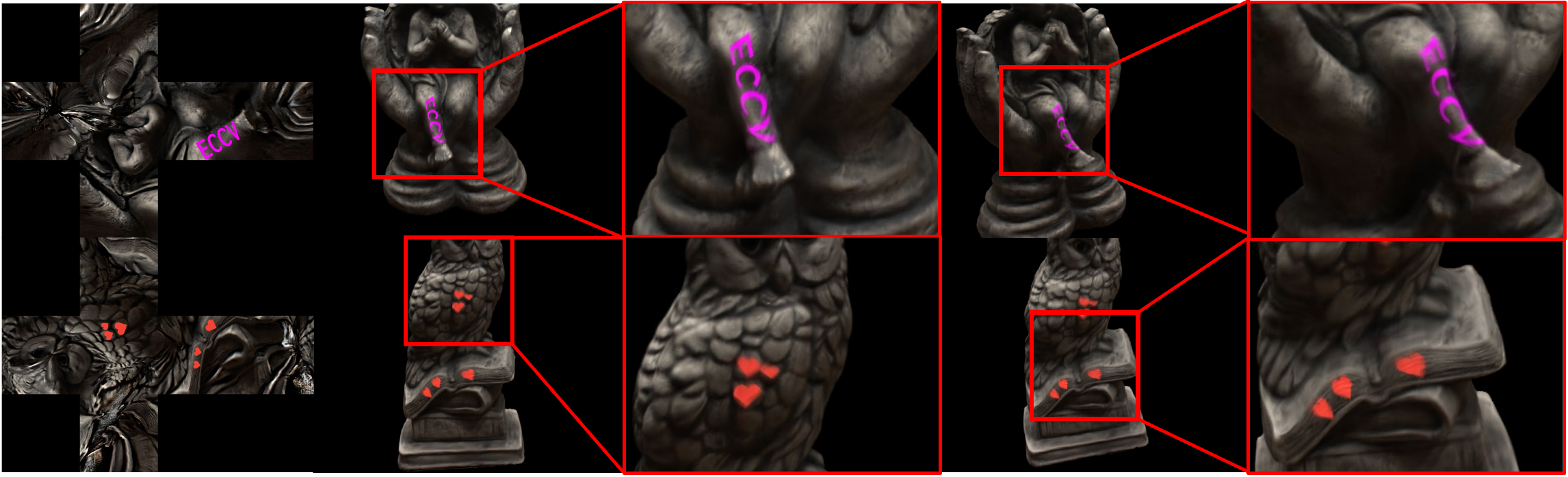}
  
\vspace{-0.3cm}
  \caption{Visualization of texture painting results of our method}
  
\vspace{-0.6cm}
  \label{fig:tex_edit}
\end{figure}

\noindent\textbf{Texture Painting.} We also investigate the capability of our method in fine-grained texture modification and show the visual results in \cref{fig:tex_edit}. We conduct fine-grained editing on the reconstructed texture by adding the ECCV text on the leg of stone statue and painting hearts on the wing. Notably, even for small patterns such as hearts, our method can yield consistent and photo-realistic rendering results across multiple views, demonstrating the potential of our method in practical 3D design workflows.

\subsection{Ablation Study}

\begin{table}[tb]
  
  \vspace{-0.5cm}
  \caption{Ablation studies of our method on the DTU dataset}
  \vspace{-0.5cm}
  \label{tab:ablate}
  \centering
    \setlength{\tabcolsep}{2pt}
    \begin{subtable}[t]{0.5\linewidth}
    \caption{Different components}
    \vspace{-0.3cm}
    \centering
    \label{tab:ablate_comp}
        \resizebox{0.9\linewidth}{!}{
            \begin{tabular}{c|ccc}
            \toprule
            \multirow{2}{*}{Method} & \multicolumn{3}{c}{DTU} \\
            & PSNR↑ & L1↓ & LPIPS↓\\
            \midrule
            Ours & 30.03 & 0.0135 & 0.1440 \\
            Ours(no SH) & 27.63 & 0.0187 & 0.1566\\
            w.o. Reg & 30.62 & 0.0126 & 0.1374 \\
            w.o. Reg(no SH) & 25.10 & 0.0264 & 0.1757 \\
            Pre-fetching & 29.28 & 0.0149 & 0.1557\\
            w.o. Pruning & 30.39 & 0.0129 & 0.1310 \\
            \bottomrule
            \end{tabular}
        }
        % \vspace{-0.5cm}
        \end{subtable}
    \begin{subtable}[t]{0.49\linewidth}
    \caption{Spherical harmonics for texture}
    \centering
    \vspace{-0.3cm}
    \label{tab:ablate_shtex}
\resizebox{0.9\linewidth}{!}{
  \begin{tabular}{c|cccc}
    \toprule
    \multirow{2}{*}{SH Deg} & \multicolumn{4}{c}{DTU Scene 114} \\
     & \#Params & FPS & PSNR↑ & L1↓ \\
    \midrule
    D=0 & 18.6 M & 55 & 24.64 & 0.0305 \\
    D=1 & 72.6 M & 31 & 24.79 & 0.0302 \\
    D=2	& 162.6 M & 14 & 25.43 & 0.0280 \\
    D=3 & 288.6 M & 7 & 27.88 & 0.0188 \\
    Ours & 20.8 M & 47 & 27.80 & 0.0189 \\
  \bottomrule
  \end{tabular}
  }
    \end{subtable}
\end{table}

We conduct ablation studies on the DTU dataset to evaluate the effectiveness of the different components and reports the quantitative results in \cref{tab:ablate}.

\begin{figure}[tb]
  \centering
  \includegraphics[width=\linewidth]{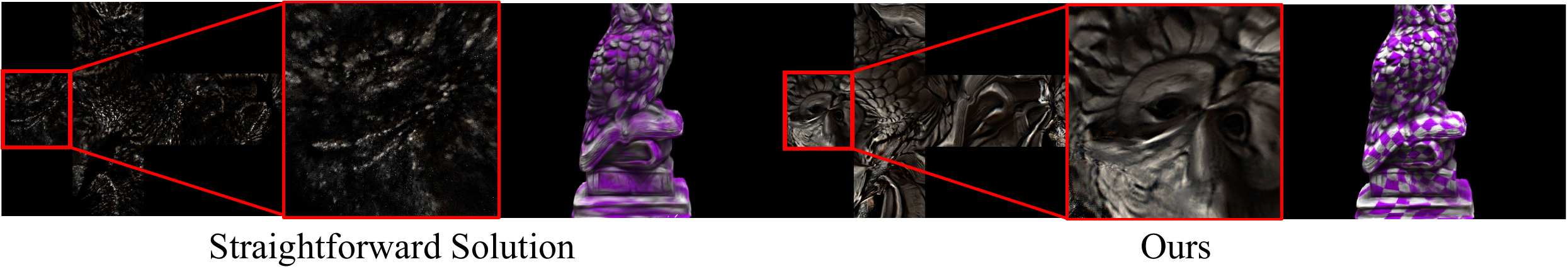}
  \vspace{-0.7cm}
  \caption{Ablation study of intersection-based UV mapping.}
  \vspace{-0.3cm}
  \label{fig:pre-fetching}
\end{figure}

\noindent\textbf{Intersection-based UV mapping.} We compare our method with the straightforward solution (Pre-fetching), which pre-fetches per-Gaussian colors from the textures using the centers before rendering. As shown in \cref{tab:ablate}, since our method enhances the representation power of each Gaussian by enabling distinct colors for different pixels, our method achieves a consistent improvement of all metrics. We also provide a visual analysis of the reconstructed textures in \cref{fig:pre-fetching}. Our method can reconstruct continuous and high-quality textures, whereas the texture space of pre-fetching is discontinuous due to lack of supervision around the mapped centers in the UV space. The view synthesis results after changing the texture with a chessboard further demonstrate the importance of intersection-based UV mapping.

\begin{figure}[tb]
  \centering
  % \vspace{-0.3cm}
  \includegraphics[width=\linewidth]{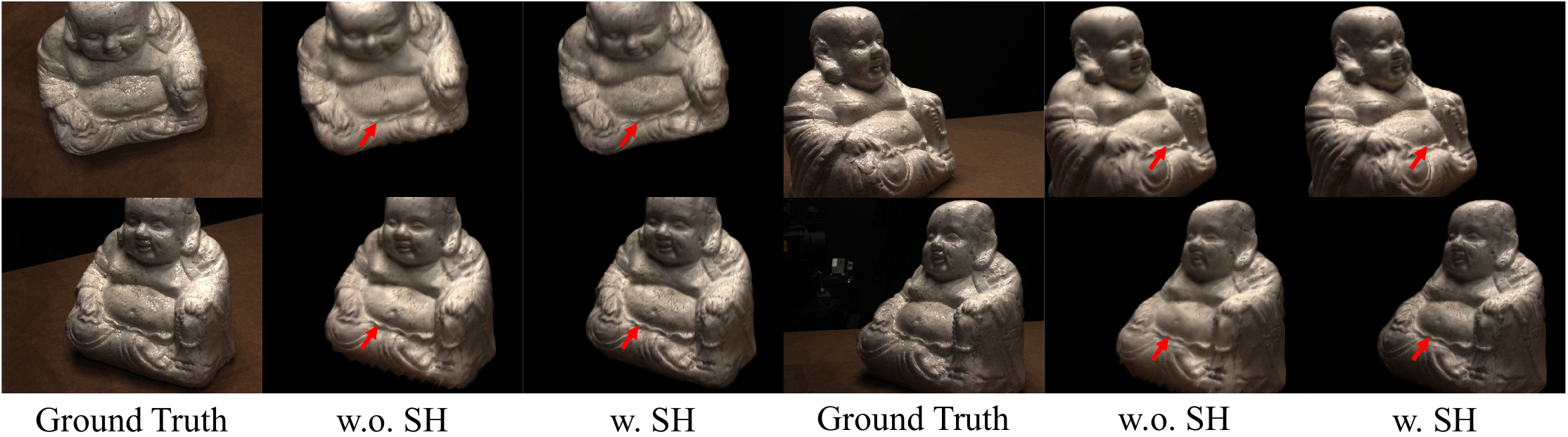}
  \vspace{-0.6cm}
  \caption{Ablation study of the per-Gaussian SH coefficients on the DTU scene 114}
  \vspace{-0.6cm}
  \label{fig:ablate_sh}
\end{figure}

\noindent\textbf{Per-Gaussian SH Coefficients.} We also utilize the trained model to render test images without per-Gaussian term $c_j^\text{SH}$ in \cref{eq:hybrid_func}. As shown in \cref{tab:ablate_comp}, although eliminating the SH coefficients (Ours(no SH)) leads to a performance decline in terms of average metrics (PSNR 30.03 → 27.63), our method can still yield photo-realistic view synthesis results. \cref{fig:ablate_sh} shows the visual comparison on DTU scene 114, which contains a porcelain material statue. We observe that the per-Gaussian SH coefficients mainly capture the view-dependent appearance, such as darkening the belly region, while the 2D texture represents the detailed view-independent appearance.

\noindent\textbf{Per-Gaussian SH v.s. SH-based Texture.} An alternative approach to capturing view-dependent appearance is to represent the texture image with a grid of SH coefficients. \cref{tab:ablate_shtex} illustrates the model complexity, rendering speed and visual quality of textures with different SH degrees on DTU scene 114. Textures with SH coefficients enhance the representation power to capture complex appearance variations under different views, albeit at the expense of substantial memory usage and computational cost. Conversely, our method applies SH coefficients with 3 degrees on the Gaussian attributes, thereby achieving high visual quality with negligible computational overhead and additional memory usage. 

\noindent\textbf{$\mathcal{L}_\text{1}^\text{noSH}$ and $\mathcal{L}_\text{ssim}^\text{noSH}$ regularization.} To validate the effectiveness of no SH coefficients rendering regularization, we present comparison results in \cref{tab:ablate_comp}. Although the ablation of regularization term (w.o. Reg) yields a minor improvement of average metrics (PSNR 30.03 → 30.62), it also leads to a significant performance decline for view synthesis without per-Gaussian SH coefficients (w.o. Reg (noSH)). The results demonstrate that without the regularization, more appearance information of reconstructed objects is captured by per-Gaussian attributes, which may hinder downstream editing applications.

\begin{figure}[tb]
  \centering
  \includegraphics[width=\linewidth]{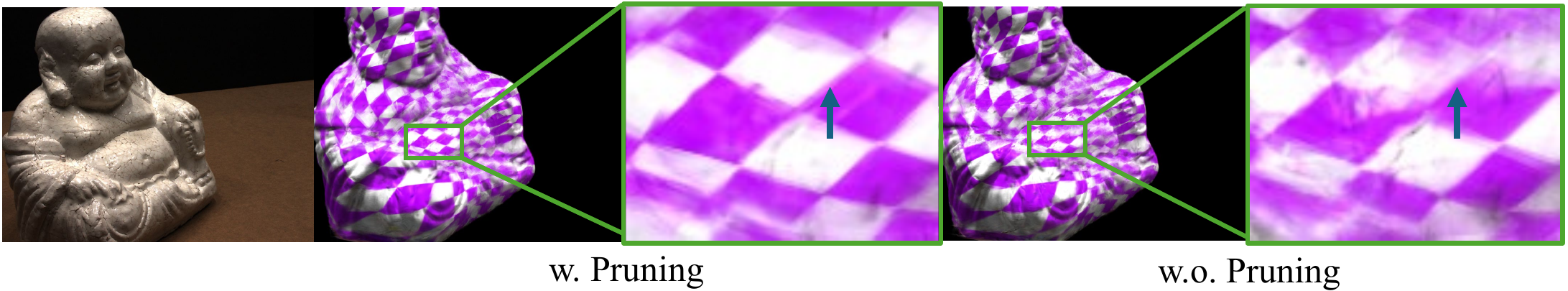}
  \vspace{-0.6cm}
  \caption{Ablation study of the pruning strategy during 3D Gaussian optimization}
  \vspace{-0.5cm}
  \label{fig:ablate_prune}
\end{figure}

\noindent\textbf{Pruning Strategy.} We employ a pruning strategy and opacity regularization to remove 3D Gaussians with high transparency. \cref{tab:ablate_comp} shows that ablating the pruning strategy (w.o. Pruning) only leads to a minor improvement in average metrics, while increasing the geometry complexity and reducing the rendering speed (31 FPS v.s. 58 FPS). \cref{fig:ablate_prune} shows the visual comparison for texture swapping. The pruning strategy can also deblur the view synthesis results for high-contrast texture swapping, such as the chessboard.  

\subsection{Limitations}

As shown in \cref{fig:ablate_prune}, the edges of the chessboard texture are not distinctly clear. We argue that the blurring issue primarily stems from the inaccurate orientations of 3D Gaussians, which leads to incorrect UV coordinates for color fetching. In addition, we define the texture space as a unit sphere, which is ill-suited to represent multiple objects or outdoor scenes. Representing the UV mapping with multiple charts, such as Nuvo~\cite{nuvo}, can address this problem, which is not the focus in this paper.   
\section{Conclusion}

We present a novel method, namely Texture-GS, which disentangles the appearance and geometry for 3D-GS to facilitate various appearance editing operations, such as texture swapping. To achieve this, we incorporate a texture mapping module into 3D-GS, which consists of a UV mapping MLP that projects 3D points into 2D UV space, a local Taylor expansion for efficient UV mapping and a learnable 2D texture to capture the appearance of 3D scenes. Experiments demonstrate our method is not only capable of reconstructing high-fidelity textures from multi-view images, but also enables various real-time texture editing application.  We hope that this work will provide researchers with a more profound comprehension of the relationship between 3D Gaussians and meshes, serving as a launchpad for further exploration.

\section*{Acknowledgement}

Tian-Xing Xu completed this work during his internship at Tencent AI Lab. The project was supported by the Tencent Rhino-Bird Elite Training Program and the Tsinghua-Tencent Joint Laboratory for Internet Innovation Technology. 

\bibliographystyle{splncs04}
\bibliography{main}
\clearpage
\section{Implementation Details}

\noindent \textbf{Training Details.} We introduce a multi-stage training pipeline for reconstructing the geometry, UV mapping MLP, and a high-quality texture from multi-view images captured from real-world scenes. Initially, we train vanilla 3D Gaussians using a combination of the losses proposed in Sec. 4.3 to obtain an initial geometry, written as:
\begin{equation}
    \mathcal{L}_\text{GS} = \mathcal{L}_1 +  \mathcal{L}_\text{mask} + \lambda_\text{ssim} \mathcal{L}_\text{ssim} + \lambda_\text{01}  \mathcal{L}_{01} + \lambda_\text{n} (\mathcal{L}_\text{norm} + \mathcal{L}_\text{sm}).
    \label{eq:total}
\end{equation}
Here, $\lambda_\text{ssim}=0.2$, $\lambda_\text{01}=0.001$ and $\lambda_\text{n}=0.1$ are used to balance the loss components. Following 3D-GS~\cite{3dgs}, we initialize the parameters of 3D Gaussians with the point clouds produced by Structure-from-Motion (SfM)~\cite{schoenberger2016sfm} techniques and optimize them for 30K iterations. To ensure the stability of the optimization process, we only apply the additional constraints $\mathcal{L}_{01}$, $\mathcal{L}_\text{norm}$ and $\mathcal{L}_\text{sm}$ after 2K iterations. We prune any semi-transparent 3D Gaussians with opacity values less than 0.5 every 6K iterations. Furthermore, we flatten each 3D Gaussian along its shortest axis by resetting the scaling value of the shortest axis as $e^{-20}$ 
%%%YKL I think you meant 1e-20 rather than e^-20 (where e is 2.718...)
% It is e^-20, here an activation layer exp is adopted to ensure the positive of scale value, we reset the value before activation as -20
every 1K iterations. Subsequently, we freeze the parameters of 3D Gaussians and render depth maps for training the UV mapping MLP, utilizing the loss function $\mathcal{L}_\text{UV}$ introduced in Sec. 4.2. Finally, we reconstruct a high-quality texture from multi-view images and finetune the parameters of 3D Gaussians using differentiable texture mapping-based splatting. Owing to the highly under-constrained nature of the parameter space for jointly optimizing the position of 3D Gaussians, the UV mapping MLP, and the texture, we freeze the UV mapping MLP and optimize the remaining parts. We supervise them with the loss function defined in Sec. 4.3, written as:
\begin{equation}
    \mathcal{L} = \mathcal{L}_\text{GS} + \lambda(\mathcal{L}_\text{1}^\text{noSH} + \lambda_\text{ssim} \mathcal{L}_\text{ssim}^\text{noSH}),
\end{equation}
where the hyperparameter $\lambda =2$ is used to encourage most (especially view-independent) appearance information to be stored in the texture. We start by optimizing only the texture image for 10K iterations, followed by jointly learning the parameters of 3D Gaussians and the texture for another 30K iterations. 

\noindent \textbf{Network Architectures.} We train a UV mapping network $\phi$, which regresses a 2D UV coordinate $u\in \mathbb{R}^2$ for each 3D point $x \in \mathbb{R}^3$ either on or close to the underlying surface of the scene, and couple it with an inverse 3D-to-2D network $\phi^{-1}$. We use Multi-Layer-Perceptrons (MLPs) to learn the functions $\phi$ and $\phi^{-1}$, which consists of 4 linear layers with feature dimensions set as 128. Notably, we only employ multi-resolution hash encoding~\cite{instantngp} on the inverse function $\phi^{-1}$ to ensure the local smoothness of the UV mapping $\phi$, similar to NeuTex~\cite{neutex}.

\begin{figure}[tb]
  \centering
  \includegraphics[width=\linewidth]{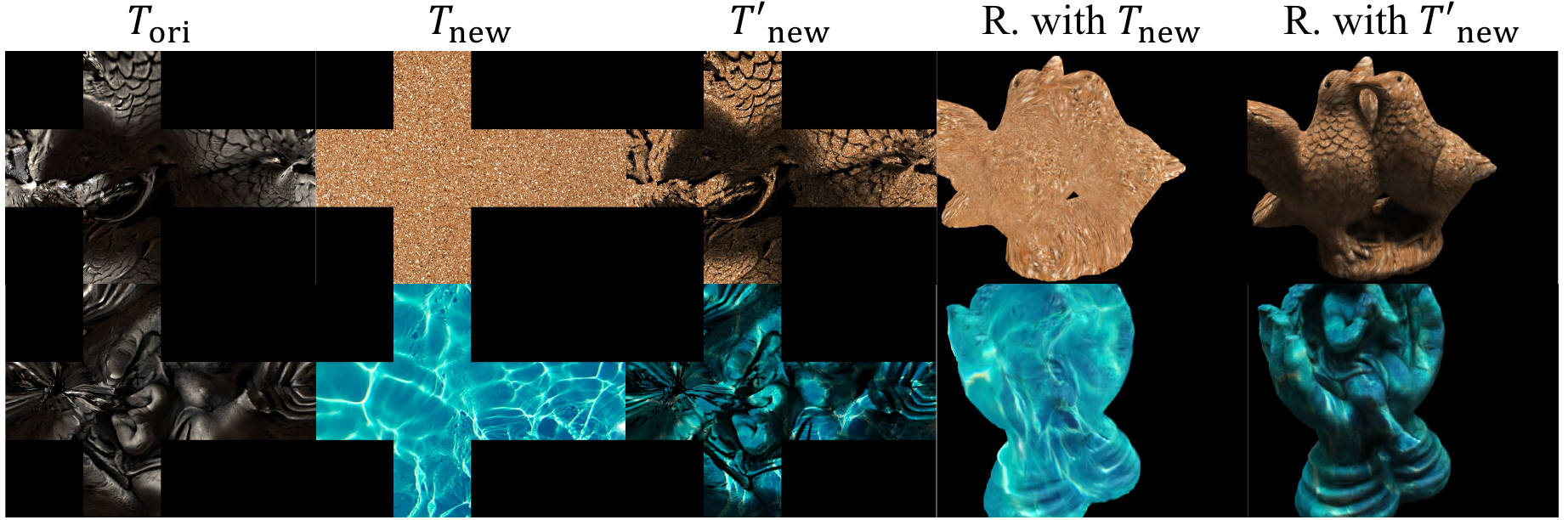}
  \vspace{-0.4cm}
  \caption{Shadow-preserving texture swapping. }
  \label{fig:sp_texture}

\vspace{-0.2cm}
\end{figure}

\noindent \textbf{Shadow-preserving Texture Swapping.} We employ a 2D texture to capture the view-independent appearance of the 3D scene from multi-view images. Notably, the appearance consists of the base color and ambient occlusion of the scene. To maintain the ambient occlusion during texture swapping, we follow NeuTex~\cite{neutex} and apply an ambient mask to the new texture, namely shadow-preserving texture swapping. Technically, let us denote the reconstructed texture and the new texture as $\mathcal{T}_\text{ori} \in \mathbb{R}^{H\times W \times 3}$  and $\mathcal{T}_\text{new} \in \mathbb{R}^{H \times W \times 3}$, respectively. We replace $\mathcal{T}_\text{ori}$ with an altered version of $\mathcal{T}_\text{new}$, denoted as $\mathcal{T}_\text{new}'\in \mathbb{R}^{H\times W \times 3}$, which is given by:
\begin{equation}
    \mathcal{T}_\text{new}' = \mathcal{T}_\text{new} \times \text{mean}(\text{min}(\mathcal{T}_\text{ori} \times 3, 1.0)),
\end{equation}
where $\text{mean}(\cdot)$ is the mean operation along the channel axis. ~\cref{fig:sp_texture} shows the visual comparison between directly swapping with the texture image $\mathcal{T}_\text{new}$ and our method. Leveraging the ambient mask, our method can preserve the shadows present in the original texture, thereby achieving photo-realistic results.

\section{More Analysis}

\begin{figure}[tb]
  \centering
  \begin{subfigure} {\linewidth}
  \centering
  \includegraphics[width=0.9\linewidth]{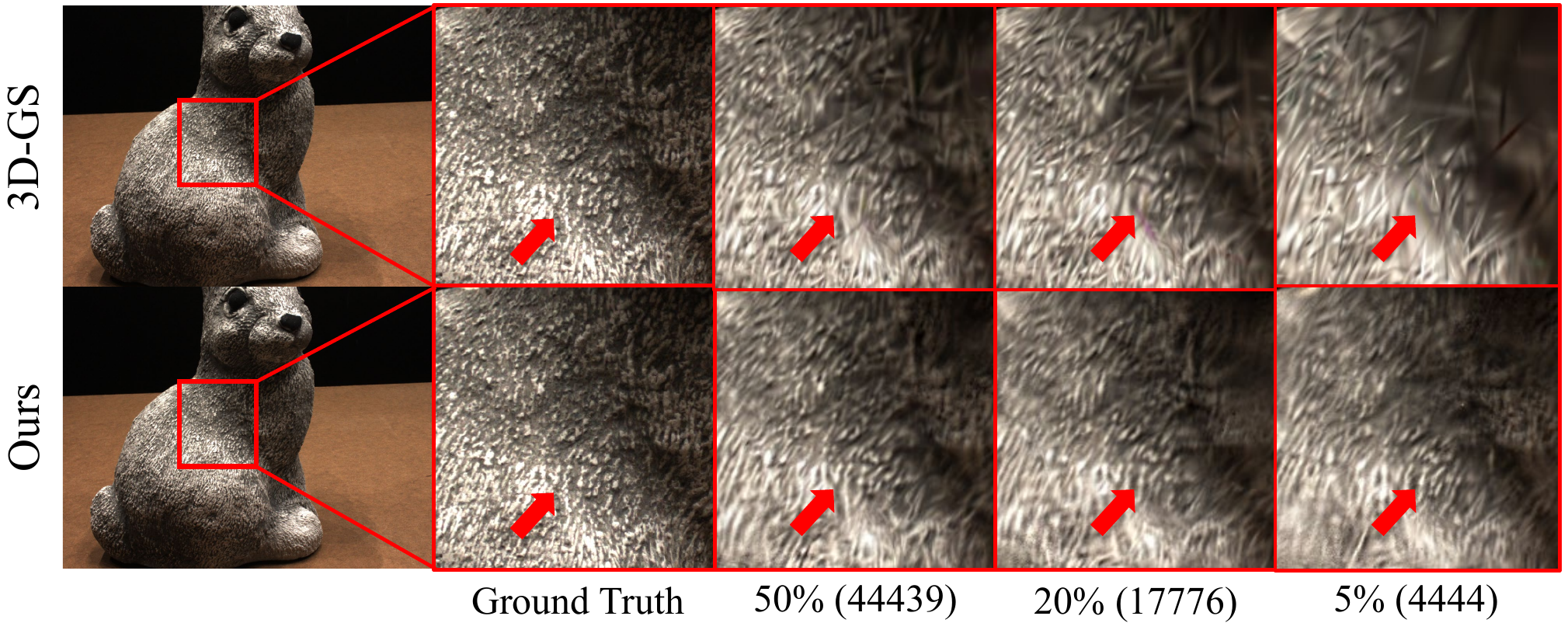}
  % \vspace{-0.2cm}
  % \caption{Under different numbers of 3D Gaussians}
  % \label{fig:tex_swapping}
  \end{subfigure}
  \vspace{-0.3cm}
    \begin{subfigure} {\linewidth}
      \centering
      \includegraphics[width=\linewidth]{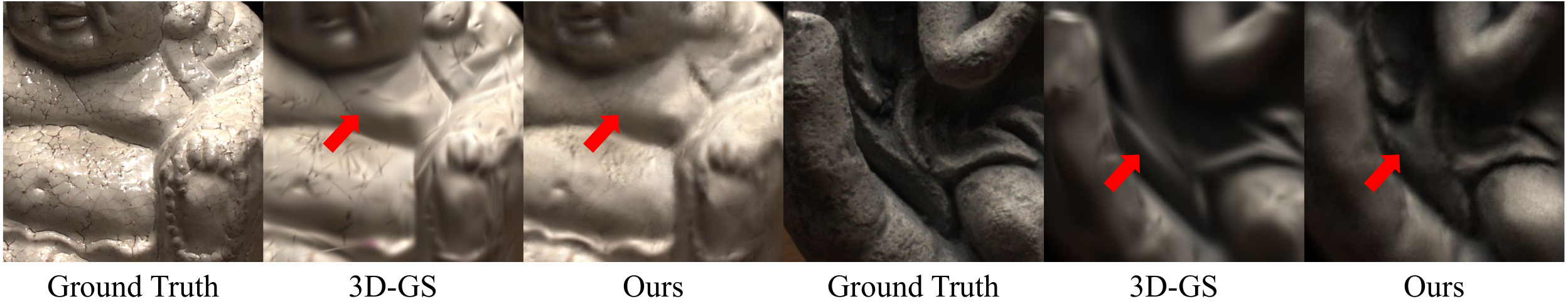}
      % \vspace{-0.2cm}
      % \caption{Different scenes}
    \end{subfigure}
    
% \vspace{-0.2cm}
  \caption{Visual comparison with 3D-GS under the same number of 3D Gaussians. Please zoom-in for a better view.}
    \label{fig:visual_comp}
  
\vspace{-0.5cm}
\end{figure}

\noindent \textbf{Number of Gaussians.} We present a visual comparison between our method and the vanilla 3D Gaussians under the same number of Gaussians to evaluate the improvement in representation power. As shown in \cref{fig:visual_comp}, reducing the number of 3D Gaussians leads to a minor degradation in visual quality for scenes with rich textures when using our method, whereas 3D-GS~\cite{3dgs} exhibits a significant loss of appearance details. These findings demonstrate that our proposed method significantly enhances the representational capabilities of each 3D Gaussian, ultimately improving the extensibility of 3D Gaussians to various computing platforms.

\begin{figure}[tb]
    \centering
    \includegraphics[width=\linewidth]{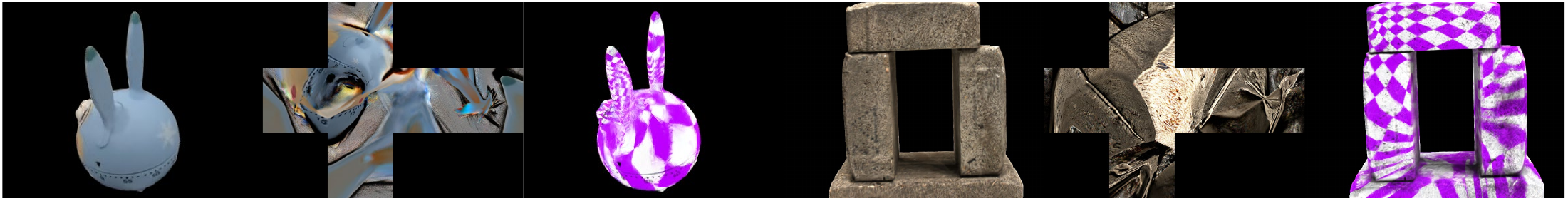}
    \vspace{-0.4cm}
    \caption{Failure cases. Due to the limited representational power of the UV mapping MLP, our method fails to learn a uniform and reasonable texture space for objects that have thin plates or holes, thereby hindering downstream applications.}
    \label{fig:limitations}
\vspace{-0.5cm}
\end{figure}

\noindent \textbf{Limitations.} The visual quality of appearance editing results relies heavily on the precision of UV mapping, which is dependent on the ray-Gaussian intersections described in Sec. 4.3 and the learned UV mapping MLP. The computation of the intersections depends on the normal vectors of 3D Gaussians, which are supervised by the pseudo ground truth normal maps derived from SfM~\cite{schoenberger2016sfm} results under the local planarity assumption. Consequently, for objects with complex geometry that do not satisfy the assumption, our method may produce sub-optimal view synthesis results for texture swapping. Besides, we eliminate the positional encoding from the UV mapping MLP to ensure local smoothness, which inevitably restricts the representation capability of the MLP. Furthermore, we define the 2D UV space as a unit spherical domain, which poses challenges when dealing with 3D scenes containing multiple objects, thin plates or holes, as shown in Fig.~\ref{fig:limitations}. In such cases, the UV mapping MLP struggles to generate accurate 2D coordinates, potentially hindering downstream applications. Representing the 2D UV space with multiple charts, such as Nuvo~\cite{nuvo}, offers a potential solution to address the challenges posed by complex geometries and multiple objects. However, multiple charts can also lead to discontinuities at the boundaries between charts, which may ultimately confound the global appearance editing operation such as texture swapping.

\begin{figure}[tb]
    \centering
    \begin{subfigure} {\linewidth}
        \includegraphics[width=\linewidth]{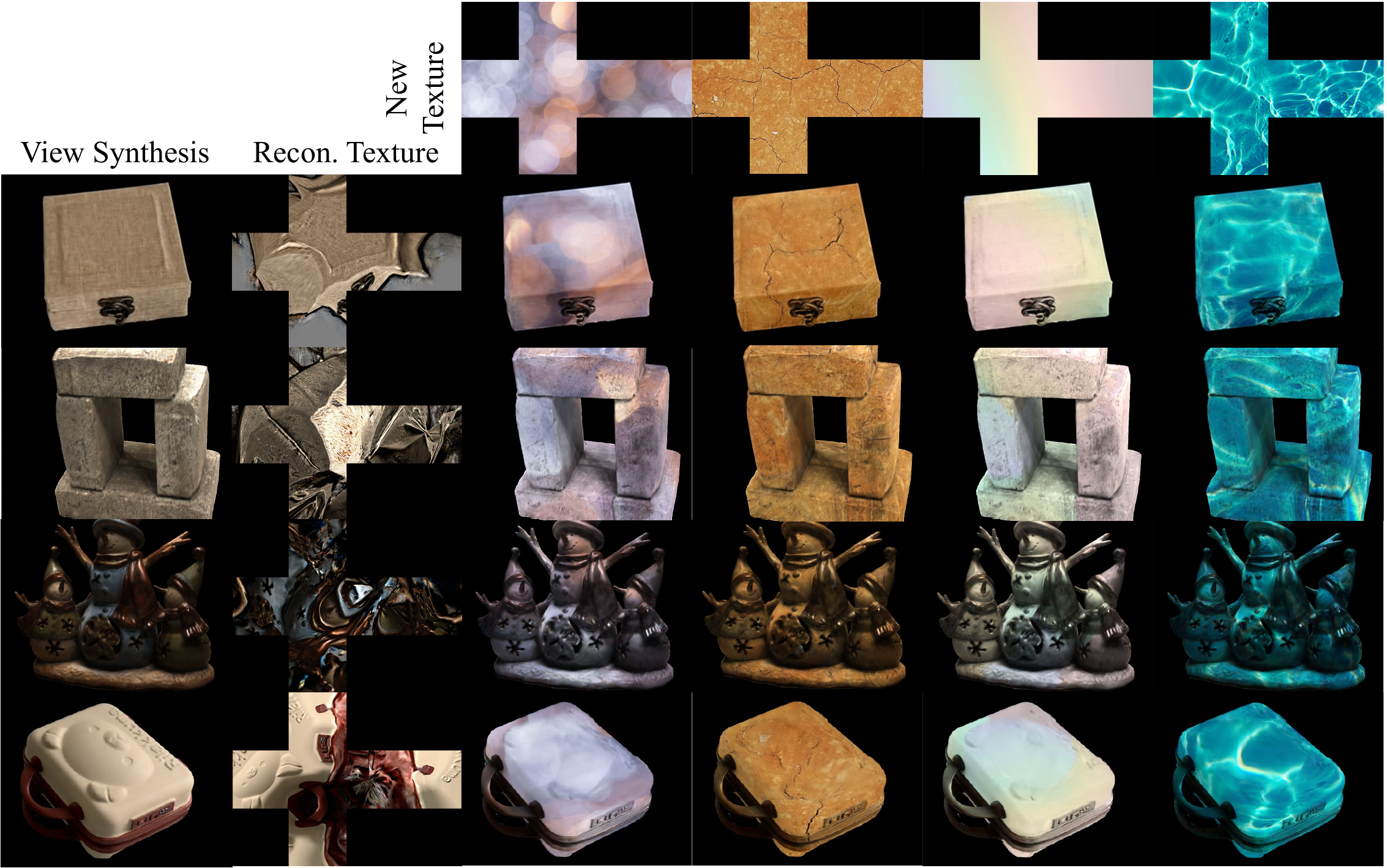}
        \vspace{-0.4cm}
        \caption{Under various textures}
    \end{subfigure}
    % \vspace{-0.2cm}
    \begin{subfigure} {\linewidth}
        \centering
        \includegraphics[width=\linewidth]{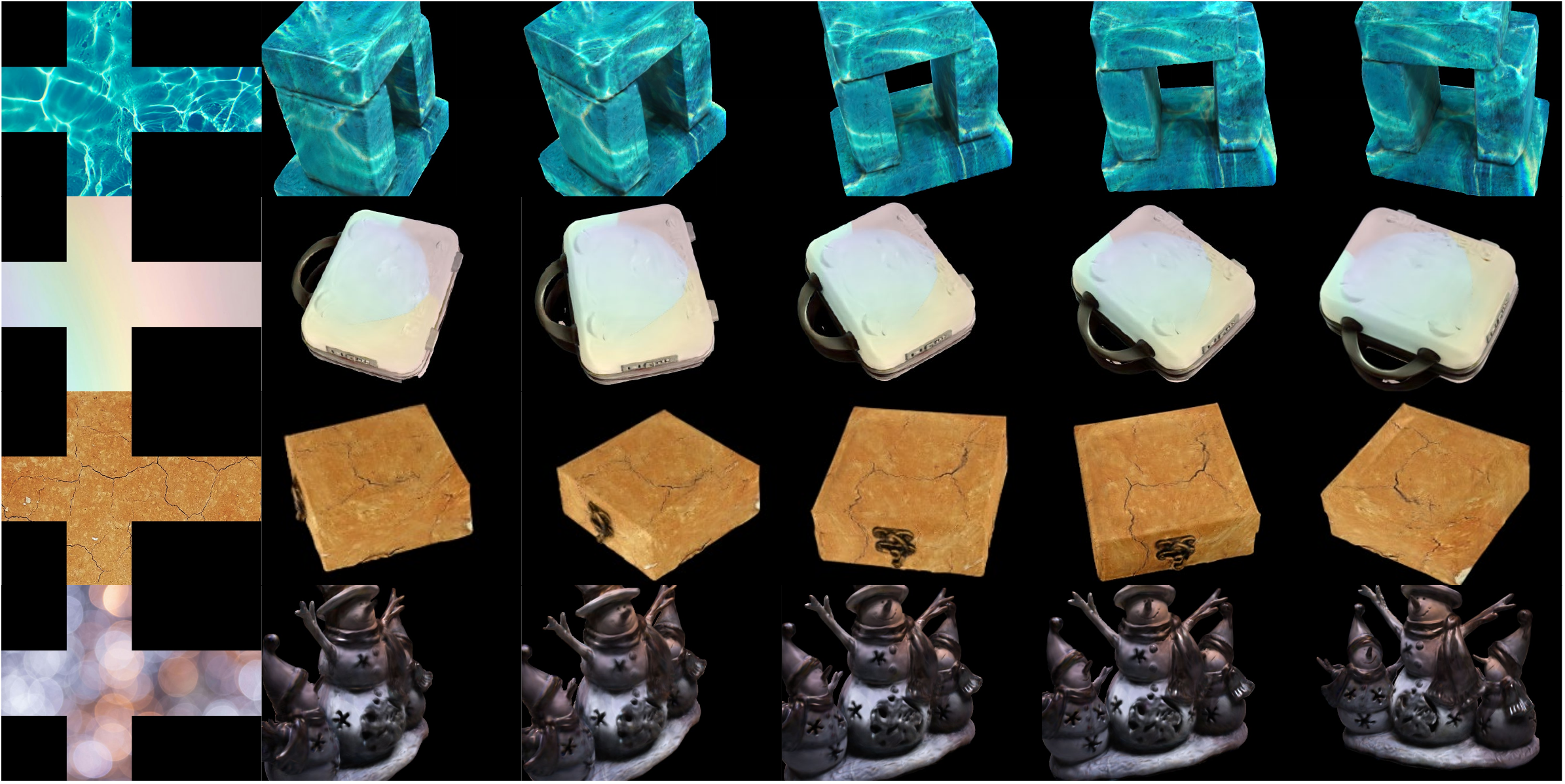}
        \vspace{-0.4cm}
        \caption{Under various views}
    \end{subfigure}
    \vspace{-0.2cm}
    \caption{More visual results for texture swapping with Texture-GS.}
    \label{fig:visual_results}
\vspace{-0.5cm}
\end{figure}

\noindent \textbf{More Visual Results.} As illustrated in Figure~\ref{fig:visual_results}, we present additional qualitative results of texture swapping on real-world scenes from the DTU dataset~\cite{dtu} and the Omni3D dataset~\cite{brazil2023omni3d}. These results demonstrate the ability of our approach to generate photo-realistic images under various textures while maintaining a consistent appearance across different viewpoints.

\end{document}